\pdfoutput=1
\documentclass[11pt]{article}

\usepackage[final]{acl}

\usepackage{times}
\usepackage{latexsym}

\usepackage[T1]{fontenc}
\usepackage[utf8]{inputenc}

\usepackage{microtype}

\usepackage{inconsolata}

\usepackage{graphicx}

\usepackage{colortbl}

\usepackage{array} % 引入 array 包
\usepackage{ulem}

\usepackage{subfiles}
\usepackage{amsmath}

\usepackage{booktabs}
\usepackage{multirow}
\usepackage{graphicx}

\usepackage{pgfplots}
\pgfplotsset{compat=1.18} % 版本可能需要根据你的安装调整
\usepackage{subcaption}

\usepackage{algorithm}
\usepackage{algpseudocode}

\usepackage{amssymb}
\usepackage{ stmaryrd }

\usepackage{enumitem}

\usepackage{pgfplots}
\pgfplotsset{width=0.48\linewidth, compat=1.15}

\title{Investigating Inference-time Scaling for Chain of Multi-modal Thought:\\A Preliminary Study}

\author{
 \textbf{Yujie Lin\textsuperscript{1,3}}\thanks{These authors contributed equally.},
 \textbf{Ante Wang\textsuperscript{1,3}}\footnotemark[1],
 \textbf{Moye Chen\textsuperscript{2}},
 \textbf{Jingyao Liu\textsuperscript{1}},
 \textbf{Hao Liu\textsuperscript{2}}
 \\
 \textbf{Jinsong Su\textsuperscript{1,3,4}}\thanks{Corresponding author.} and
 \textbf{Xinyan Xiao\textsuperscript{2}}
\\
 \textsuperscript{1}School of Informatics, Xiamen University, China
 ~~\textsuperscript{2}Baidu Inc., Beijing, China \\
 \textsuperscript{3}Key Laboratory of Digital Protection and Intelligent Processing of Intangible Cultural Heritage\\ of Fujian and Taiwan (Xiamen University), Ministry of Culture and Tourism, China \\
 \textsuperscript{4}Shanghai Artificial Intelligence Laboratory
\\
\texttt{\small{\{yjlin,wangante\}@stu.xmu.edu.cn}}~~ \texttt{\small{jssu@xmu.edu.cn}}
}

\begin{document}
\maketitle
\begin{abstract}

Recently, inference-time scaling of chain-of-thought (CoT) has been demonstrated as a promising approach for addressing multi-modal reasoning tasks.
While existing studies have predominantly centered on text-based thinking, the integration of both visual and textual modalities within the reasoning process remains unexplored.
In this study, we pioneer the exploration of inference-time scaling with multi-modal thought, aiming to bridge this gap.
To provide a comprehensive analysis, we systematically investigate popular sampling-based and tree search-based inference-time scaling methods on 10 challenging tasks spanning various domains.
Besides, we uniformly adopt a consistency-enhanced verifier to ensure effective guidance for both methods across different thought paradigms.
Results show that multi-modal thought promotes better performance against conventional text-only thought, and blending the two types of thought fosters more diverse thinking.
Despite these advantages, multi-modal thoughts necessitate higher token consumption for processing richer visual inputs, which raises concerns in practical applications.
We hope that our findings on the merits and drawbacks of this research line will inspire future works in the field.\footnote{Our code can be found at \url{https://github.com/DeepLearnXMU/TTS_COMT}}

\end{abstract}

\section{Introduction}

% Wang, recent progress of inference-time scaling methods for this field
The remarkable performance of OpenAI's o1 in tackling reasoning tasks has sparked widespread research interest in studying inference-time scaling across various communities~\cite{Towards_reasoning_era}.
This technique is built upon the success of Chain-of-Thought~(CoT, \citealt{cot}), which enables large language models~(LLMs) to solve a problem through step-by-step reasoning.
It makes use of additional computation at inference time to conduct ``\textit{slow thinking}''~\cite{kahneman2011thinking} to further improve the accuracy of responses.
Taking the rich experience of pioneering research in language-only approaches~\cite{tts-mcts5,tot,scaling,tts-bs3, tts-mcts2, tts-mcts4, tree-search3, wang2024litesearch, tree-search2, wang2025don,bi2025cotkineticstheoreticalmodelingassessing}, rapid progress has been made in addressing multi-modal reasoning tasks leveraging typical methods such as Best-of-N, Beam Search, and Monte Carlo Tree Search~(MCTS)~\cite{xu2024llava,yao2024mulberry}.

\begin{figure}[t]
  \includegraphics[width=\columnwidth]{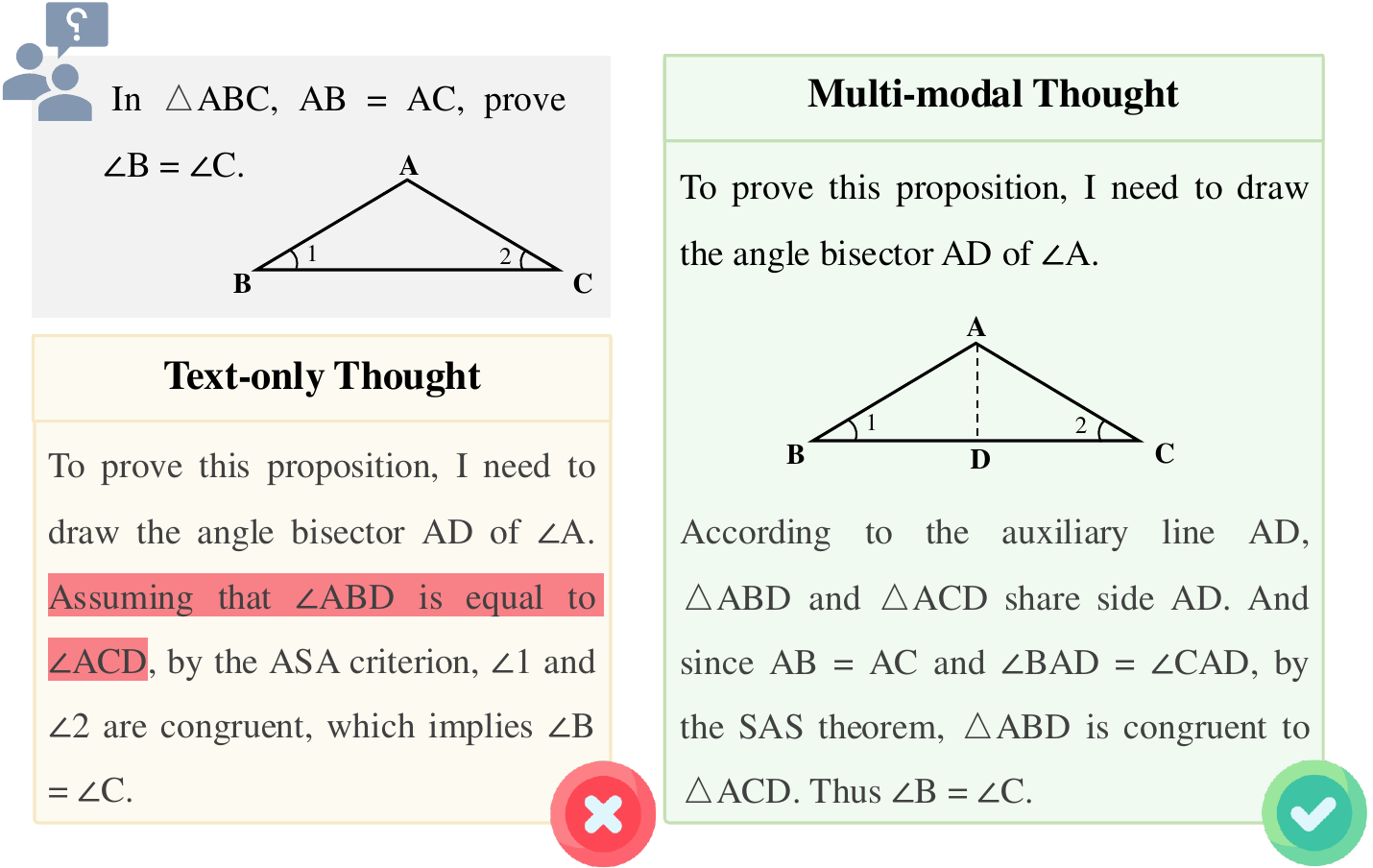}
  \caption{Solving the problem using text-only thought vs. multi-modal thought. Additional visual information from the latter offers richer and more intuitive features, making it easier to yield better subsequent steps.}
  \label{fig:cot-mcot}
\end{figure}

% Text-based CoT and Multi-modal CoT, compare them using a figure
Although effective, these methods mostly follow the practice of text-based thinking, overlooking the crucial role of thinking in the visual modality in multi-modal scenarios. This oversight leads to suboptimal performance, thus highlighting the need for new approaches that integrate both modalities.
% Taking the geometric problem in Figure~\ref{fig:cot-mcot} as an example, reasoning based on added auxiliary lines can effectively assist the model in performing correct proof.
Consider the process of solving geometric problems as an example: humans often utilize auxiliary lines to facilitate problem-solving. Analogously, rich and intuitive visual information, e.g., drawn auxiliary lines, can enhance the model's ability to generate improved solutions, as shown in Figure~\ref{fig:cot-mcot}.
Recent studies~\cite{Learn_to_explain, Mathvista, image-of-thought, vsk, Mathverse, chen2024m3cot,comt} have also demonstrated the benefits of incorporating supplementary visual information at each reasoning step compared with conventional text-based methods~\cite{mcot,ddcot,ccot}.
Therefore, an important yet under-explored question arises: \textit{What is the potential of inference-time scaling of multi-modal thinking?}

% This study, sampling-based and tree search-based, process verifier
To this end, we conduct the first study on this problem, shedding light on the merits and constraints of this new research line.
We follow \citet{cot} and \citet{vsk} to elicit text-only and multi-modal thought, respectively.
For inference-time scaling, we investigate popular sampling-based and tree search-based methods.
Sampling-based method generates multiple independent samples in parallel, applying techniques like Self-Consistency~\cite{tts2-Self-Consistency} or Best-of-N~\cite{tts-mcts1, tts-mcts3} to select the most reliable reasoning trails.
However, this approach is inefficient because it requires exploring full solution paths, even if a mistake has occurred early.
Tree search-based method tackles this issue by utilizing sophisticated search algorithms, such as Beam Search~\cite{tot} and MCTS~\cite{tts-mcts7}.
Both these methods ask for a critic to discriminate the promising samples or search steps.
We consistently employ an advanced Large Vision Language Model~(LVLM) as a verifier to infer critic scores and, alternatively, enhance this method by aggregating results from multiple trials.
In this way, it effectively provides reliable feedback without the need for specific training.

% Experiment, main conclusion*, and discussion of future direction
We conduct a comparison of inference-time scaling using multi-modal and text-only thought across 10 datasets, encompassing tasks related to geometric reasoning, mathematical reasoning, and visual question answering.
The main findings are as follows:
\begin{itemize}[leftmargin=12pt]
    \item The use of multi-modal thought achieves significantly better performance and higher upper bounds compared to text-only thought on average, demonstrating the potential of this line of research.
    \item Although effective, it requires higher token consumption to process richer visual inputs. This result calls for more efficient reasoning mechanisms, especially for practical concerns.
    \item Further analyses show that tree search-based methods effectively mitigate reasoning errors, such as invalid processed images. However, their success heavily relies on verifier performance, highlighting the need to develop more effective multi-modal verifiers.
\end{itemize}
% 需要想些其他的分析，围绕scaling（性能上限、性能涨幅和多样性）和critic（多模态CoT更容易判断还是更难），这两点才是inference-time scaling关注的
We hope that these findings can inspire future research to develop more advanced and robust methods for multi-modal reasoning.

\section{Preliminaries}

In this section, we firstly establish the formulation for both text-only and multi-modal thought~(\textsection\ref{cot}), and then provide a detailed explanation of two inference-time scaling paradigms~(\textsection\ref{scaling}): sampling-based and tree search-based methods.
In addition, we present a prompting-based verifier equipped with a consistency-enhanced mechanism to effectively guide these inference-time scaling methods~(\textsection\ref{verifier}).

\subsection{Thought Formulation} \label{cot}
Chain-of-Thought (CoT, \citealt{cot}) is a technique that encourages the model to generate intermediate reasoning steps $\mathbf{S}=(\mathbf{s}_1,...,\mathbf{s}_n)$ before reaching the final answer $\mathbf{a}$.
In the context of multi-modal reasoning tasks, given a problem $\mathbf{q}$ and images $\mathbf{I}$, each reasoning step $\mathbf{s}_i$ is sampled from an LVLM $\mathcal{M}$:
\begin{equation}
\label{eq:cot}
    \mathbf{s}_i \sim \mathcal{M}\big(\mathbf{q}, \mathbf{I}, \mathbf{s}_{1:i-1}\big).
\end{equation}
This expansion allows for more computations to be deployed and can unlock the ability to solve more complex problems~\cite{lichain}.

Following the success on language-only reasoning tasks, most previous works~\cite{mcot, ddcot, ccot} study \textit{text-only thought}, where $\mathbf{S}$ is defined as a text string.
In this work, we aim to explore the potential of \textit{multi-modal thought}, where any reasoning step $\mathbf{s}_i$ can blend multi-modal information, e.g., both text and image.

% 这也可以在discussion中讲
However, as most prevalent LVLMs still struggle to generate reliable images, we follow \citet{vsk} to allow the LVLM to generate executable code for visual manipulation instead.
Implementation details can be found in Appendix~\ref{sec:app_vsk}.

\subsection{Inference-Time Scaling} \label{scaling}

\subsubsection{Sampling-based Methods}
These methods scale inference-time computation by generating multiple reasoning chains in parallel, and then determining the most promising answer from them. Here we investigate two widely used techniques, Self-Consistency~\cite{tts2-Self-Consistency} and Best-of-N~\cite{tts-mcts1, tts-mcts3}.

\paragraph{Self-Consistency}
This method leverages the intuition that the correct answer typically can be reached using multiple different ways of thinking~\cite{tts2-Self-Consistency}.
Given $N$ sampled reasoning chains $\mathbf{S}_1,...,\mathbf{S}_N$ generated from the LVLM, it selects the most consistent answer via majority voting over their corresponding answers $\mathbf{a}_1,...,\mathbf{a}_N$:
\begin{equation}
    \mathbf{a}^* = \arg\max_{\mathbf{a} \in \{\mathbf{a}_1,...,\mathbf{a}_N\}} \sum_{i=1}^{N} \mathbb{I}(\mathbf{a}_i = \mathbf{a}),
\end{equation}
where $\mathbb{I}(\cdot)$ denotes the indicator function and $\mathbf{a}^*$ represents the selected answer.

\paragraph{Best-of-N}
This approach utilizes trained verifiers to evaluate the correctness of model generated solutions.
Compared to Self-Consistency based on voting, it can be more efficient when a strong verifier is available.
Formally, given $N$ sampled reasoning chains and a verifier $\mathcal{F}(\cdot)$, we have
\begin{equation}
    \mathbf{S}^*=\max_{\mathbf{S}\in \{\mathbf{S}_1,...,\mathbf{S}_N\} }\mathcal{F}(\mathbf{S}),
\end{equation}
where $\mathbf{S}^*$ is the reasoning chain with the highest verifier score, from which the final answer $\mathbf{a}^*$ is derived.

\subsubsection{Tree Search-based Methods}
Sampling-based methods are easy to implement due to their simplicity.
However, they are inefficient because they require exploring full solution paths, even if a mistake occurs early~\cite{xiang2025towards}.
To address this challenge, tree search-based methods are later explored.
They model the problem-solving as a tree search process, where each node represents a reasoning step.
Here, we investigate the most prevalent Beam Search~\cite{tts-bs1, tts-bs2, tot, tts-bs3} and MCTS~\cite{tts-mcts5,tts-mcts8,  tts-mcts2, tts-mcts4, tts-mcts6, tts-mcts7, yao2024mulberry} algorithms in both language-only and multi-modal tasks.

\paragraph{Beam Search}

Adapted from classical beam search algorithms used in sequence generation tasks (e.g., machine translation~\cite{beamsearch4mt, beamsearch4mt2}), this method dynamically explores and prunes reasoning paths at each step for deductive reasoning.
The algorithm operates with a beam width \( B \) (number of retained reasoning chains per step) and an expansion size \( N \) (new thoughts generated per candidate).
Starting with \( N \) initial candidate thoughts sampled from a question $q$ via Equation~\ref{eq:cot}, a verifier \( \mathcal{F}(\cdot) \) scores all candidates, retaining only the top-\( B \) chains.
Each retained chain is then expanded by sampling \( N \) new thoughts, producing \( B \times N \) candidates, after which the verifier reevaluates and prunes the pool to the top-\( B \) chains.
This cycle of scoring and pruning repeats iteratively until all active chains reach finished states (final answers), at which point the highest-scoring finished chain across all iterations is selected as the final answer \( \mathbf{a}^* \).

\paragraph{MCTS}

It is a decision-making algorithm that is powerful for solving complex problems, as demonstrated by AlphaGo~\cite{alphago, alphago1}. MCTS operates by iteratively exploring the most promising reasoning paths through four key phases: selection, expansion, evaluation, and backpropagation. 

In the selection phase, the algorithm traverses the tree from the root node, recursively choosing child nodes using the Upper Confidence Bound (UCB) metric~\cite{ucb}. This metric strategically balances the exploitation of high-value paths and the exploration of under-visited paths. The UCB formula is defined as:  

\begin{equation}
    \text{UCB}(\mathbf{s}_i) = \frac{V(\mathbf{s}_i)}{C(\mathbf{s}_i)} + w \times \sqrt{\frac{\ln C(\mathbf{s}_{i-1})}{C(\mathbf{s}_i)}},
\end{equation}
where \(V(\mathbf{s}_i)\) and \( C(\mathbf{s}_i)\) denotes the accumulated value and count of being visited of \(\mathbf{s}_i\). \(w\) is a weighting parameter that adjusts the exploration-exploitation balance.
During expansion, the algorithm extends the selected node by generating a new thought through Equation~\ref{eq:cot}.
The evaluation phase then estimates the value of this newly expanded node using a verifier \(\mathcal{F}(\cdot)\).
Finally, backpropagation updates the values of all ancestors, propagating the evaluation results back to the root.

After multiple iterations, the finished reasoning chain with the highest accumulated value is selected, yielding the optimal final answer \(\mathbf{a}^*\).

\begin{figure}[htp]
  \includegraphics[width=\columnwidth]{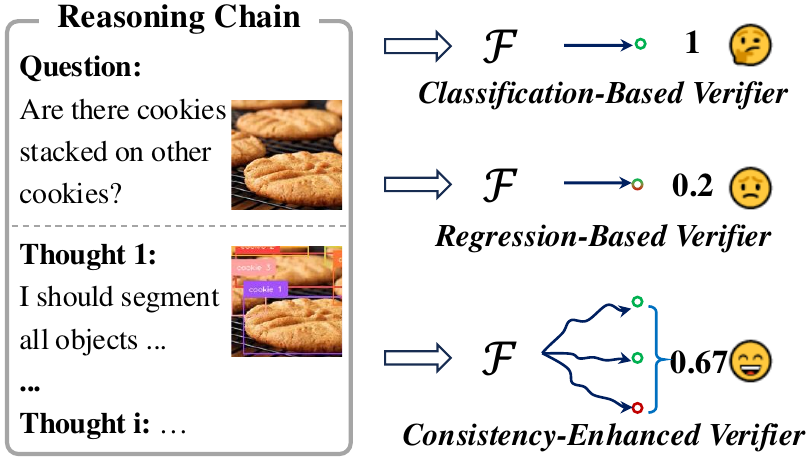}
  \centering
  % \caption{\textbf{Consistency-Enhanced Verifier:} The reasoning chain is formatted as input to the LVLM, which is then instructed to verify it through CoT reasoning. By sampling multiple verification responses, the score is then computed by aggregating the verification results.}
  \caption{Three types of verifiers investigated in this work, where the classification-based verifier outputs sparse binary scores (0 or 1) and the regression-based verifier provides dense but inaccurate scores. We introduce the consistency-enhanced verifier to compute dense and accurate scores by aggregating multiple evaluations sampled from the classification-based verifier.}
  \label{fig:verifier}
\end{figure}

\subsection{Verifiers for Inference-Time Scaling} \label{verifier}

The verifier \(\mathcal{F}\) is pivotal in inference-time scaling, tasked with evaluating the validity of reasoning chains. 
Existing approaches fall into two categories: training-based~\cite{llava-critic, shepherd} and prompting-based~\cite{binary1, binary2, scoring1}.
Training-based approaches hold theoretical promise by optimizing verifiers on corresponding labeled data.
However, acquiring such annotations is challenging, resulting in datasets limited in scale and diversity, thereby restricting the generalizability of trained verifiers.
Thus, this work focuses on prompting-based approaches, capitalizing on the strong instruction-following capabilities of LVLMs to assess reasoning chains.  

Formally, given a reasoning chain \(\mathbf{s}_{1:i}\), the verifier \(\mathcal{F}\) operates as:  
\(\mathbf{r} = \mathcal{M}(\mathbf{P}, \mathbf{s}_{1:i})\),
where \(\mathbf{P}\) represents the verification instruction, and \(\mathbf{r}\) is the evaluation output, which includes a scalar score.
Particularly, we investigate two instruction designs for \(\mathbf{P}\): 
\begin{itemize}[leftmargin=*]
    \item \textbf{Classification-Based}~\cite{binary1, gen-verifier,  binary2}: It prompts the verifier to classify the reasoning chain as ``correct'' or ``incorrect.'' While intuitive, this binary output provides sparse feedback, offering limited granularity to guide searching. 
    \item \textbf{Regression-Based}~\cite{scoring1, llava-critic}: Here, the verifier is prompted to output a continuous score \(\in [0, 1]\) reflecting the quality of chains. However, directly predicting precise scores is error-prone, often yielding high-variance results that degrade performance.\footnote{The prompts used for these two designs are in Appendix~\ref{sec:app_verifier_inst}.}
\end{itemize}

To address these limitations, we introduce the \textbf{Consistency-Enhanced Verifier}, compared with the above two in Figure~\ref{fig:verifier}.
This method aggregates $N_{v}$ independent evaluations using classification-based instructions, computing the proportion of ``correct'' classifications across trials. 
In this way, it not only avoids variance in regression-based outputs but also overcomes the sparsity of single-trail classification, enabling more reliable guidance for inference-time scaling.

\begin{table*}[t]
\resizebox{\textwidth}{!}{%
\begin{tabular}{@{}lcccccccccc@{}}
\begin{tabular}{l|cccccccccc} 
\toprule
\multicolumn{1}{c|}{\multirow{2}{*}{Methods}} & \multicolumn{4}{c}{\textbf{Geometry}}                         & \multicolumn{3}{c}{\textbf{Math}}             & \multicolumn{3}{c}{\textbf{VQA}}               \\ 
\cmidrule(l){2-5} \cmidrule(l){6-8} \cmidrule(l){9-11}  

\multicolumn{1}{c|}{}                         & Geometry3K    & Maxflow       & Isomorphism   & Connectivity  & Convexity     & Parity        & Winner ID     & $V^*$ Bench   & MMVP          & BLINK          \\ 
\midrule

Direct                                        & 35.4          & 28.9          & 50.8          & 52.3          & 90.2          & 72.1          & 51.8          & 51.7          & 64.0          & 60.7           \\
Multimodal-CoT                                & 43.8          & 25.8          & 50.0          & 71.1          & 91.8          & 75.8          & 57.6          & 49.2          & 65.3          & 63.3           \\
DDCoT                                         & 29.2          & 25.8          & 51.6          & 60.2          & 77.3          & 52.1          & 43.2          & 47.9          & 49.3          & 64.0           \\
CCoT                                          & 25.0          & 26.6          & 35.9          & 12.5          & 87.5          & 56.5          & 39.3          & 53.4          & 54.0          & 64.4           \\

\midrule
\multicolumn{11}{l}{\textbf{\textit{Text-only Thought + Inference-time Scaling}}} \\ \midrule

CoT                            & 39.6          & 30.9          & 43.0          & 53.9          & 91.0          & 77.9          & 56.4          & 46.2          & 65.7          & 62.2           \\

\ \ +Self-Consistency                              & 37.5       & 35.9          & 51.6          & 62.5          & 93.4          & 76.0          & 57.2          & 52.1          & 62.3          & 64.6           \\
\ \ +Best-of-N                                     & 43.8         & 39.8          & 53.1          & 60.9          & 93.4          & 81.0          & 56.0          & 53.4          & 67.3          & 65.5           \\
\ \ +Beam Search                                    & 41.7          & 54.7          & 59.4          & 65.6          & 94.5          & 85.2          & \textbf{61.1} & 37.8          & 70.0          & 59.8           \\
\ \ +MCTS                                          & 45.8          & 51.6          & 62.5          & 64.8          & 94.9          & 81.0          & 58.4          & 53.4          & 66.7          & \uline{66.2}   \\

\midrule
\multicolumn{11}{l}{\textbf{\textit{Multi-modal Thought + Inference-time Scaling}}} \\ \midrule

Visual Sketchpad                         & 33.3          & 81.3          & 86.1          & 85.9          & 93.0          & 81.0          & 53.7          & 49.1          & 66.3          & 58.2           \\ 

\ \ +Self-Consistency                              & 39.6          & 85.2          & 88.3          & 87.5          & \uline{96.5}  & 83.6          & 54.1          & 51.3          & 68.3          & 61.5           \\
\ \ +Best-of-N                                     & 43.8          & 91.4          & 89.8          & \uline{91.4}          & 94.1          & 82.8          & 54.9          & 55.0          & 67.7          & 66.1           \\
\ \ +Beam Search                                    & \textbf{54.2} & \textbf{95.3} & \uline{92.2}  & \textbf{99.2} & \uline{96.5}  & \textbf{87.2} & 58.4          & \textbf{64.3} & \uline{71.3}  & 64.8           \\
\ \ +MCTS                                          & \uline{52.1}  & \uline{94.5}  & \textbf{93.0} & \textbf{99.2} & \textbf{96.9} & \uline{86.7}  & \uline{60.3}  & \uline{62.2}  & \textbf{74.0} & \textbf{68.7}  \\
\bottomrule
\end{tabular}
\end{tabular}%
}

\caption{Accuracy across 10 datasets from three domains. 
For each dataset, the best result is highlighted in \textbf{bold}, and the second-best is emphasized with \uline{underlining}.}
\label{tab:main_x}
\end{table*}

\section{Experiments}

\subsection{Setup}
\paragraph{Datasets}
We conduct experiments on 10 datasets spanning three distinct domains:
\begin{itemize}[leftmargin=*]
\item \textbf{Geometric Reasoning:} To assess geometry problem-solving capabilities, we utilize the Geometry3K dataset~\cite{geometry3k} and three tasks from IsoBench~\cite{isobench}, including Connectivity, Maxflow, and Isomorphism.
\item \textbf{Mathematical Reasoning:} We evaluate mathematical reasoning performance using the Parity, Convexity, and Winner ID tasks from IsoBench~\cite{isobench}.
\item \textbf{Visual Question Answering (VQA):} To test generalizability, we also employ three challenging VQA benchmarks: $V^*$ Bench~\cite{vstar}, MMVP~\cite{mmvp}, and BLINK~\cite{blink}.
\end{itemize}

More detailed introduction to these datasets can be found in Appendix~\ref{sec:app_dataset}.

\paragraph{Baselines}
CoT~\cite{cot} and Visual Sketchpad~\cite{vsk} are employed in our experiments to implement text-only thought and multi-modal thought, respectively. Both of them are further compared under four inference-time scaling methods: Self-Consistency, Best-of-N, Beam Search, and MCTS. Additionally, the following latest multi-modal reasoning frameworks are also compared in our main experiment, including Multimodal-CoT~\cite{mcot}, DDCoT~\cite{ddcot}, and CCoT~\cite{ccot}. Detailed baseline descriptions are provided in Appendix~\ref{sec:app_baselines}.

\paragraph{Metrics} We evaluate the performance using two metrics: \textbf{Accuracy} and \textbf{Pass@K}.
Accuracy measures whether the highest-ranked answer is correct, while Pass@K represents whether the correct answer can be recalled within $K$ answers.

\paragraph{Implementation Details} 
We employ GPT-4o-mini\footnote{gpt-4o-mini-2024-07-18} as the policy and verifier by default.
% GPT-4o-mini\footnote{gpt-4o-mini-2024-07-18} serves as both our primary policy model and the LVLM in verifier.
For all experiments, we set the temperature to 1.0 during generation.
For Self-Consistency and Best-of-N, the number of samples $N$ is set to 5.
For Beam Search, we set both the beam width $B$ and expansion size $N$ to 4.
For MCTS, we set the maximum expansion per node to 4, the maximum number of iterations to 30, and the exploration constant $w$ in UCB to 1.4.
We adopt the consistency-enhanced verifier by default and set the number of samples $N_v$ to 5.

\subsection{Multi-modal vs. Text-only Thought}
\paragraph{Multi-modal thought shows greater advantages compared with text-only thought.}

Table~\ref{tab:main_x} presents compelling experimental results that highlight the superiority of multi-modal thought over both text-only thought and the latest multi-modal reasoning frameworks, particularly when enhanced by inference-time scaling.

A closer analysis reveals notable performance variations across different domains. In geometric reasoning, multi-modal thought consistently outperforms text-only thought under single-chain reasoning, with the exception of Geometry3K, where it initially lags but surpasses text-only thought when inference-time scaling is applied. In mathematical reasoning, the advantages of multi-modal thought become even more evident. For example, while text-only thought initially excels on Winner ID due to the problem’s rich textual context, multi-modal thought progressively closes the gap under inference-time scaling. The similarly superior performance observed across VQA tasks further underscores the adaptability and effectiveness of multi-modal thought, strengthening its potential for broader applications.

\paragraph{Multi-modal thought exhibits a higher performance upper limit.}

\begin{figure}

    \centering
    \includegraphics[width=\columnwidth]{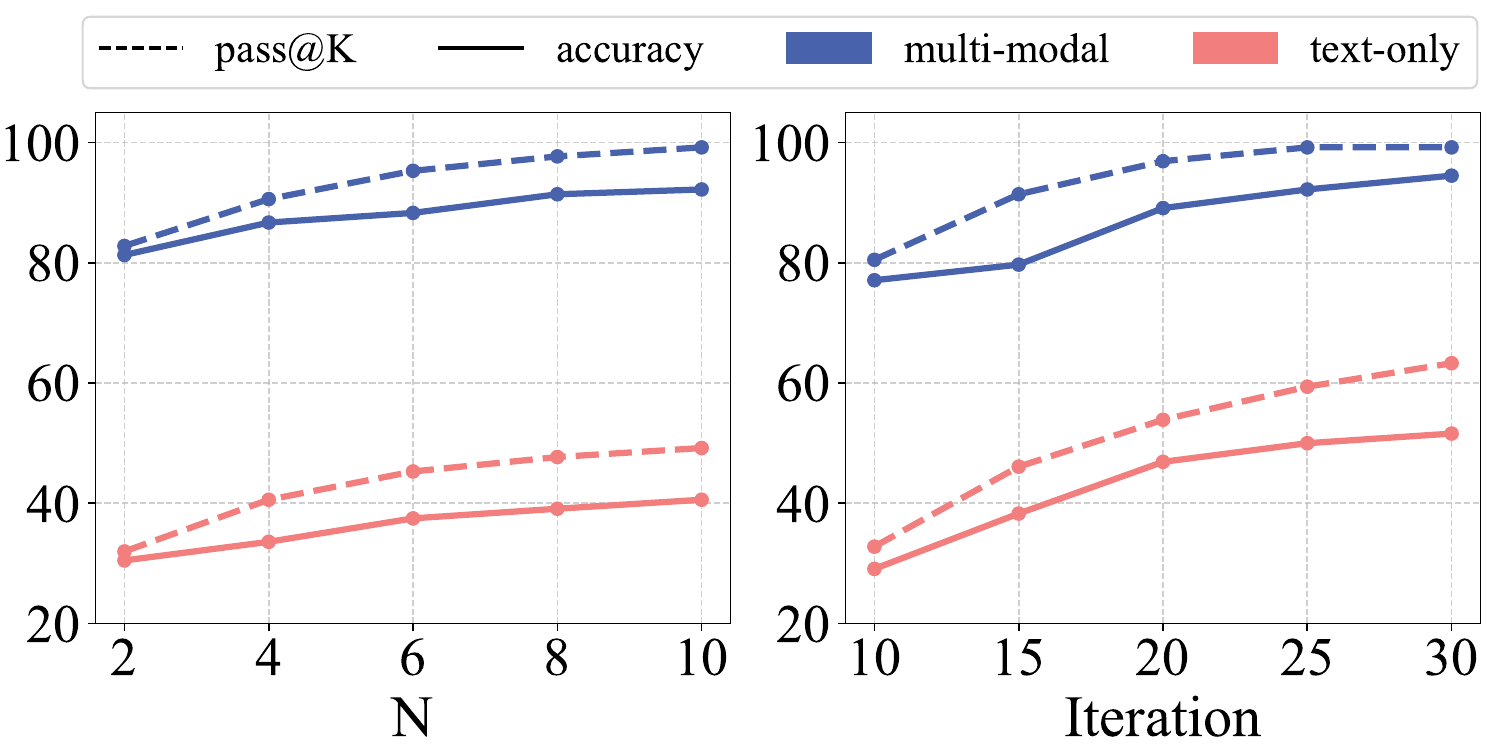}
    \caption{Performance comparison between text-only thought and multi-modal thought on the Maxflow dataset under Best-of-N (left) and MCTS (right).}
    \label{fig:oracle}
\end{figure}

To further investigate the potential of text-only and multi-modal thought, we compare their performance upper limits on the Maxflow dataset using Best-of-N and MCTS methods. As shown in Figure~\ref{fig:oracle}, within Best-of-N, accuracy consistently improves as $N$ increases, with multi-modal thought achieving a significantly higher upper limit. This trend is further validated by the MCTS results, which exhibit a similar pattern, reinforcing the superior potential of multi-modal thought.

\begin{figure}[t]
  \includegraphics[width=\columnwidth]{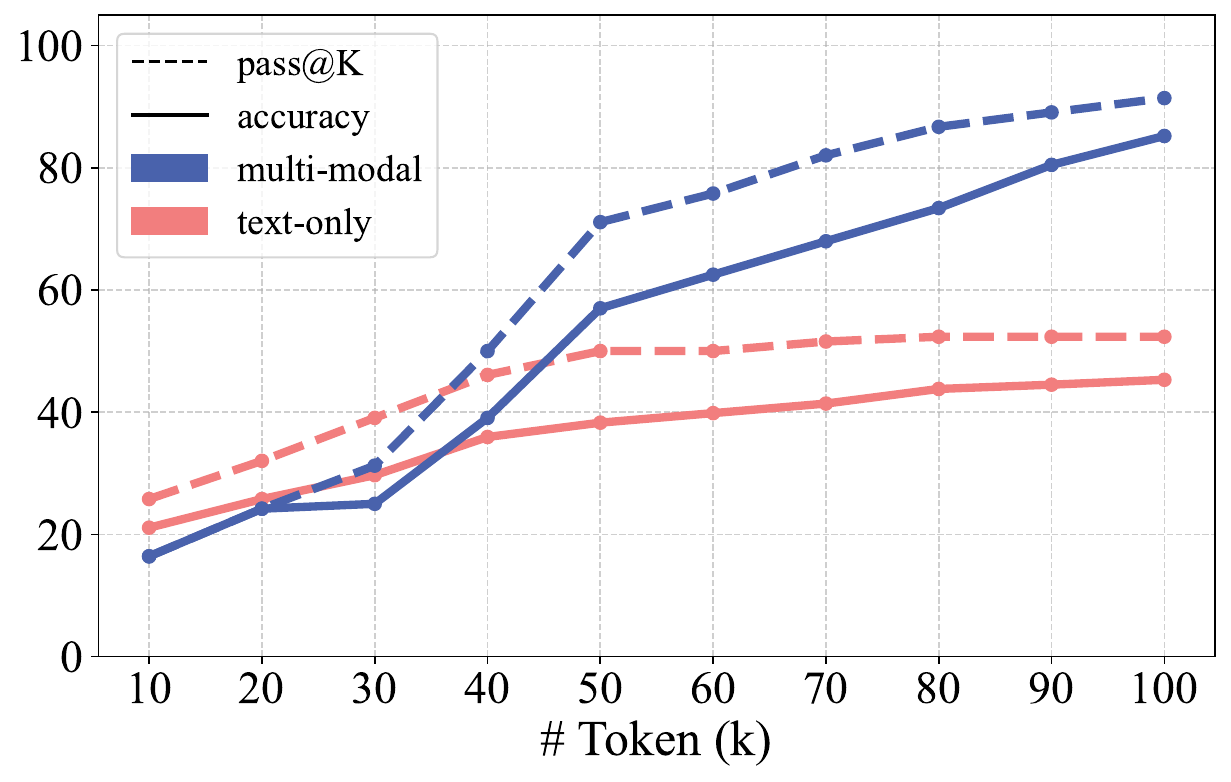}
  \caption{Performance comparison between text-only thought and multi-modal thought varies with the maximum token consumption on the Maxflow dataset under Self-Consistency.}
  \label{fig:cost}
\end{figure}

\begin{figure}[t]
\centering
  \includegraphics[width=\columnwidth]{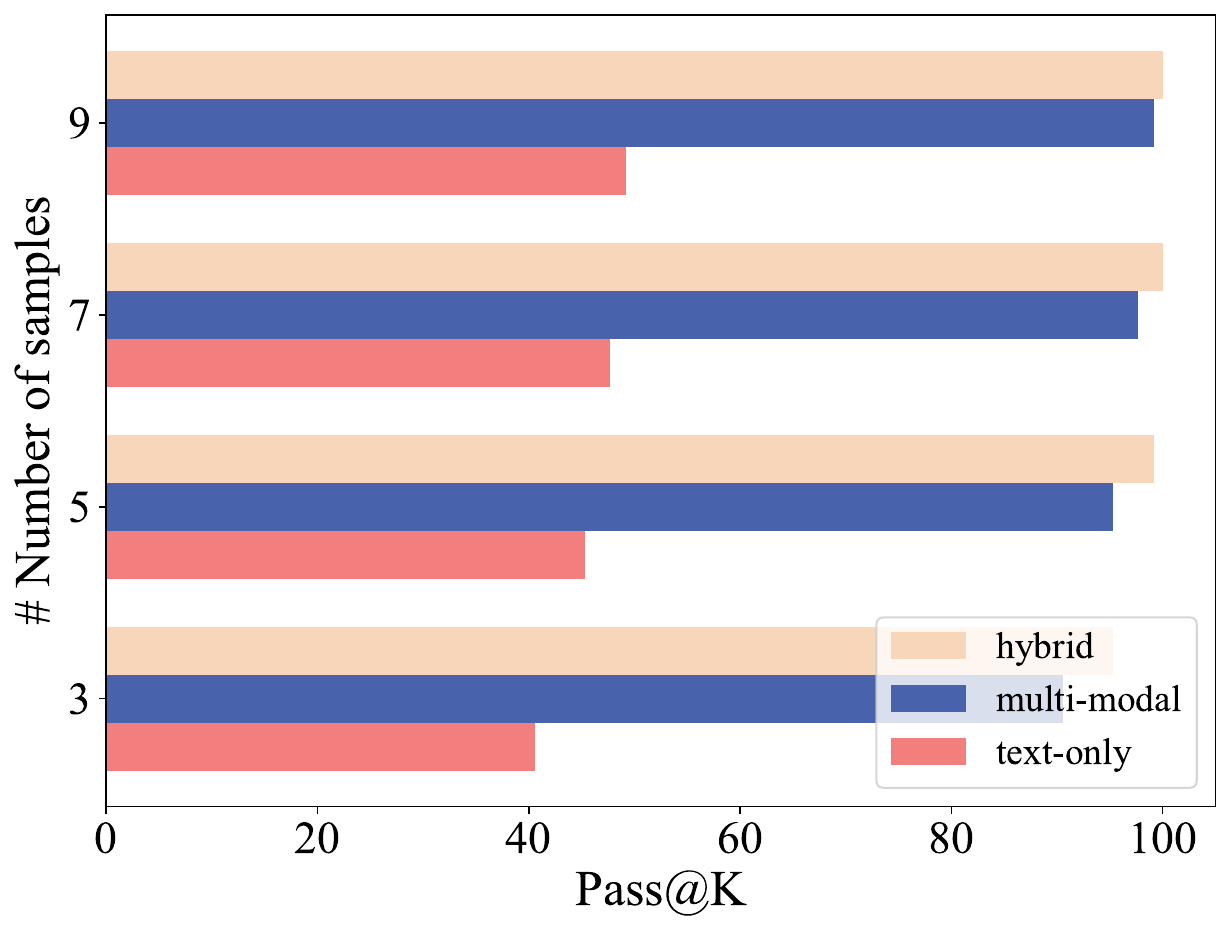}
  \caption{Performance of Self-Consistency on the Maxflow dataset as the number of samples increases, comparing text-only thought, multi-modal thought, and the hybrid form.}
  \label{fig:mix}
\end{figure}

\paragraph{The effectiveness of multi-modal thought relies on higher token consumption.} 
Given the substantial difference in token cost between processing text and image information in LVLMs, it is crucial to understand the trade-off between performance and token usage for text-only and multi-modal thought. To this end, we conduct experiments on the Maxflow dataset using Self-Consistency to analyze their performance under inference-time scaling with varying token limits per sample. 

Figure~\ref{fig:cost} illustrates the relationship between token limits and performance for both text-only and multi-modal thought. Under strict token constraints, multi-modal thought underperforms compared to text-only thought. This is likely due to the limited number of reasoning steps multi-modal thought can sample when the token budget is severely restricted, preventing the model from effectively processing visual information.

However, as the token limits are relaxed, the performance of multi-modal thought improves rapidly. With a more generous token budget, the model can fully leverage visual information, leading to significant performance gains. This highlights a key trade-off: while multi-modal thought enables more complex reasoning, it requires significantly more tokens to process visual inputs effectively. This suggests that future research should prioritize developing methods for compressing or optimizing the token consumption associated with image processing in LVLMs, enabling more efficient utilization of multi-modal thought.

\paragraph{Multi-modal and text-only thought are complementary.}

While our results demonstrate the overall effectiveness of multi-modal thought, we further examine whether text-only thought retains value in certain scenarios. We hypothesize that the two approaches may be complementary, each leveraging different aspects of the input. To test this, we conduct experiments on the Maxflow dataset using the Self-Consistency method, comparing text-only thought, multi-modal thought, and a hybrid form, i.e., voting from both text-only thought and multi-modal thought.

As presented in Figure~\ref{fig:mix}, the hybrid form outperforms both text-only and multi-modal thought individually, suggesting that text-only thought provides valuable insights that enhance reasoning, even in cases where multi-modal reasoning alone may fall short. This performance boost underscores the complementary nature of the two approaches and highlights the potential of a balanced integration of them for more robust multi-modal reasoning.

\subsection{Ablation Study on the Verifier}

\paragraph{Consistency-enhanced verifier outperforms naive prompting-based alternatives.}

\begin{figure}[t]
    \centering
  \includegraphics[width=\columnwidth]{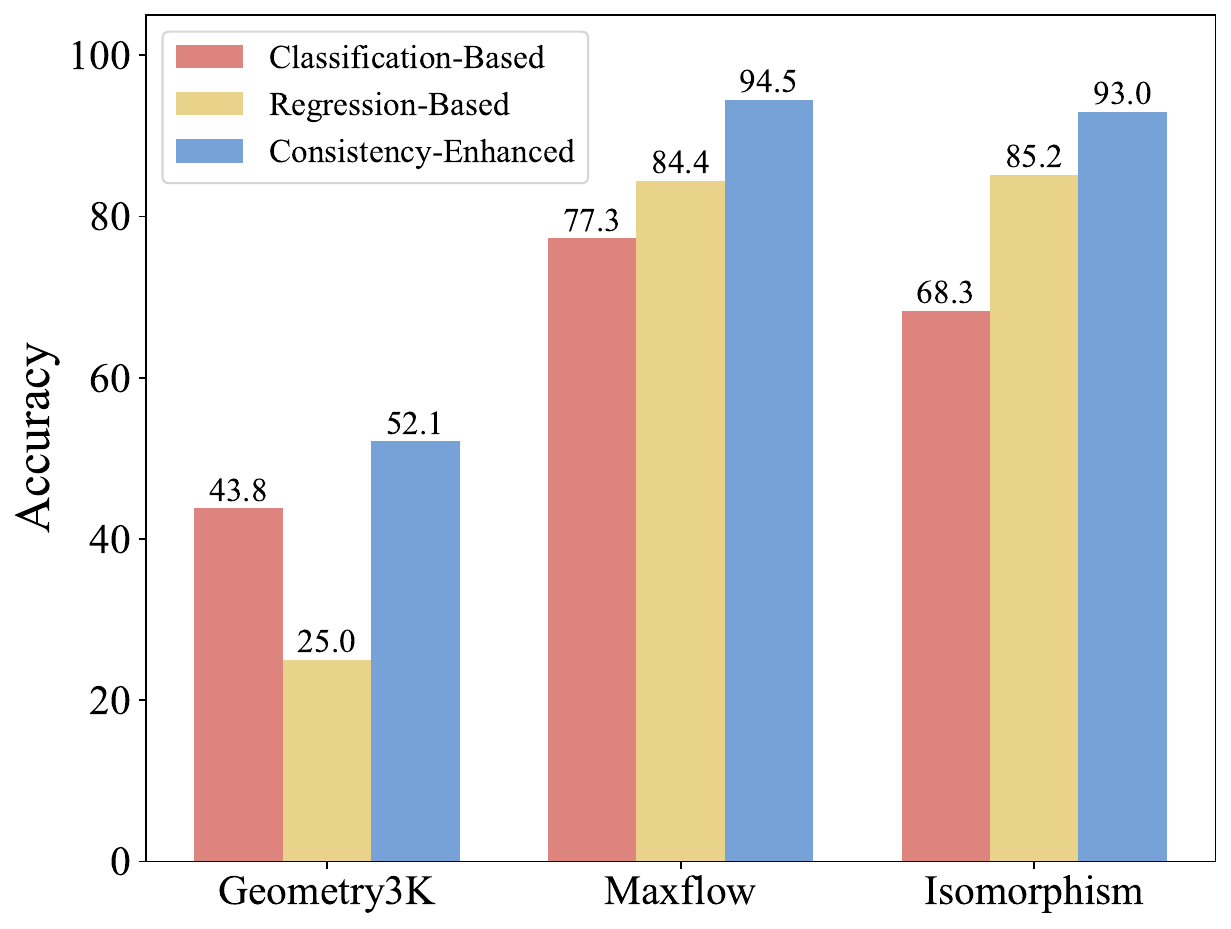}
  \caption{Performance comparison of Classification-Based, Regression-Based, and Consistency-Enhanced Verifiers using MCTS with multi-modal thought across three datasets.}
  \label{fig:verifier-com}
\end{figure}

To evaluate the superiority of our proposed consistency-enhanced verifier, we compare its performance against classification-based and regression-based verifiers, as described in~\textsection\ref{verifier}. We conduct this comparison on the Geometry3K, Maxflow, and Isomorphism datasets using MCTS with multi-modal thought. As shown in Figure~\ref{fig:verifier-com}, our verifier consistently outperforms both naive prompting-based approaches across all three datasets, demonstrating its effectiveness in addressing the limitations of existing approaches and providing more reliable guidance for inference-time scaling.

\begin{figure}[t]
  \includegraphics[width=\columnwidth]{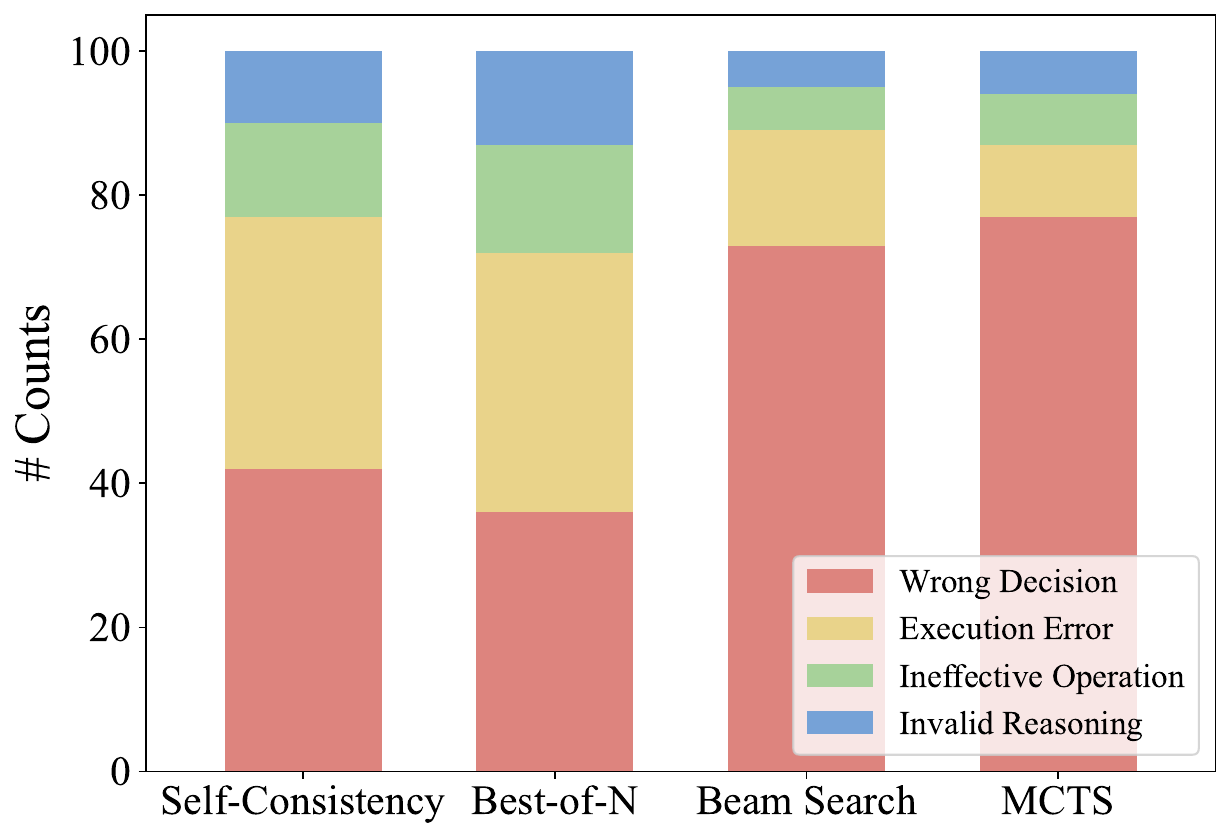}
  \caption{Error statistics on four inference-time scaling methods with multi-modal thought.}
  \label{fig:error_static}
\end{figure}

\paragraph{Consistency-enhanced verifier performance scales with $N_v$.}

Our consistency-enhanced verifier is designed based on the hypothesis that aggregating multiple verifications enhances robustness and performance. Specifically, we expect accuracy to improve as the number of sampled verifications increases. To validate this, we conduct experiments using MCTS on the Geometry3K, Maxflow, and Isomorphism datasets, systematically varying $N_v$. From the results in Table~\ref{tab:N-verifier}, we observe a clear trend: increasing $N_v$ consistently improves performance. This finding strongly supports our hypothesis, confirming the effectiveness of the design rationale for the verifier.
% \usepackage{booktabs}
% \usepackage{graphicx}
\begin{table}[t]
\resizebox{\columnwidth}{!}{%
\begin{tabular}{@{}cccc@{}}
\toprule
    & \phantom{aa}Geometry3K\phantom{aa} & \phantom{a}Maxflow\phantom{a} & \phantom{aa}Isomorphism\phantom{aa} \\ \midrule
\phantom{aaa}$N_{v}$=3\phantom{aaa} &    41.7      &   89.8       &  91.4        \\
\phantom{aaa}$N_{v}$=5\phantom{aaa} &    52.1      &   94.5       &  93.0        \\
\phantom{aaa}$N_{v}$=7\phantom{aaa} &    58.3      &   97.7       &  96.1        \\ \bottomrule
\end{tabular}%
}
\caption{Performance variants with $N_{v}$ under MCTS method with multi-modal thought.}
\label{tab:N-verifier}
\end{table}

\subsection{Error Analysis}

To gain deeper insights into the behavior of multi-modal thought under different inference-time scaling methods, we conduct a detailed error analysis. We identify four primary sources of error: Wrong Decision, Execution Error, Ineffective Operation, and Invalid Reasoning (detailed explanations are provided in Appendix~\ref{sec:app_error}) and then randomly sample 100 error cases from each method across all datasets, manually classifying the underlying causes.

\paragraph{Tree search-based methods significantly reduce intermediate errors.} \label{subsec:error_analysis}

Figure~\ref{fig:error_static} presents the distribution of error types for various inference-time scaling methods. We can find that sampling-based methods exhibit a predominance of intermediate reasoning errors~(Execution Error, Ineffective Operation, Invalid Reasoning).  Among these, Execution Error is the most prevalent, which mostly occurs in VQA tasks. This is likely due to the difficulty expert models face in accurately operating images, coupled with a lack of error-handling logic in the code.  Consequently, simple sampling methods often continue exploring flawed paths after encountering such errors, leading to the failure. In contrast, tree search-based methods significantly mitigate intermediate reasoning errors.  By incorporating verifiers at each step, these methods strategically select optimal reasoning paths, effectively preventing the exploration of erroneous reasoning steps.

\paragraph{Effective verifiers are crucial for further error reduction.}

The reduction in intermediate errors leads to a relative increase of Wrong Decision, highlighting this error type as the new primary challenge.  Further manual observation reveals that these errors primarily stem from limitations in the verifier. While effective at identifying blatant errors, the verifier struggles with more nuanced or complex reasoning steps.  This underscores the critical importance of improving verifier effectiveness, particularly in discerning subtle errors, to fully unlock the potential of multi-modal thought.

\subsection{Policy Model Impact}

\begin{table}[t]
\resizebox{\columnwidth}{!}{%
\begin{tabular}{@{}lccc@{}}
\toprule
\multicolumn{1}{l|}{Methods}            &Geometry3K\phantom{a} & \phantom{a}Maxflow\phantom{a}    & \phantom{a}Connectivity  \\ \midrule
\multicolumn{4}{l}{\textit{\textbf{Qwen2-VL-72B-Instruct}}}                            \\ \midrule
\multicolumn{1}{l|}{Direct}             & 43.8    & 10.4        & 85.4       \\
\multicolumn{1}{l|}{CoT}                & 31.3    & 35.4        & 64.6               \\
\multicolumn{1}{l|}{\ \ +Best-of-N}   & 31.3    & 35.4        & 66.7               \\

\multicolumn{1}{l|}{Visual Sketchpad}   & 41.7    & 12.5        & 93.8               \\
\multicolumn{1}{l|}{\ \ +Best-of-N}   & 45.8    & 31.3        & 98.4          \\
\midrule

\multicolumn{4}{l}{\textit{\textbf{OPENAI-gpt-4o-mini}}}                            \\ \midrule
\multicolumn{1}{l|}{Direct}             & 35.4    & 28.9        & 52.3       \\
\multicolumn{1}{l|}{CoT}                & 39.6    & 30.9        & 53.9               \\
\multicolumn{1}{l|}{\ \ +Best-of-N}   & 43.8    & 39.8        & 60.9               \\
\multicolumn{1}{l|}{Visual Sketchpad}   & 33.3    & 81.3        & 85.9               \\
\multicolumn{1}{l|}{\ \ +Best-of-N}   & 43.8    & 91.4        & 91.4               \\
\midrule

\multicolumn{4}{l}{\textit{\textbf{OPENAI-gpt-4o}}}                            \\ \midrule
\multicolumn{1}{l|}{Direct}             & 52.1    & 50.0        & 71.1       \\
\multicolumn{1}{l|}{CoT}                & 54.2    & 55.5        & 76.6               \\
\multicolumn{1}{l|}{\ \ +Best-of-N}   & 70.8    & 67.2        & 90.6               \\
\multicolumn{1}{l|}{Visual Sketchpad}   & 62.5    & 93.8        & 98.4               \\
\multicolumn{1}{l|}{\ \ +Best-of-N}   & 77.1    & 96.9        & 99.2               \\
\bottomrule
\end{tabular}%
}
\caption{Accuracy on three datasets by employing different policy models.}
\label{tab:policy}
\vspace{-10pt}
\end{table}

To assess the impact of the underlying policy model on the effectiveness of multi-modal thought, we extended our experiments beyond the previously used GPT-4o-mini to include a representative open-source LVLM Qwen2-VL-72B-Instruct~\cite{qwen2vl} and a more advanced closed-source LVLM GPT-4o\footnote{gpt-4o-2024-08-06}. The experiments are conducted on the Geometry3K, Maxflow, and Connectivity datasets, using Best-of-N as the inference-time scaling method.

\paragraph{Performance of multi-modal thought generalizes across models.}
As shown in Table~\ref{tab:policy}, multi-modal reasoning consistently improves performance across all three evaluated LVLMs, suggesting that the benefits of multi-modal thought are not model-specific but rather a generalizable advantage across different architectures and training paradigms.

\paragraph{The stronger the model, the greater the performance gains.}
An intriguing trend also emerges from the results, i.e., the performance gains from multi-modal thought correlate with the strength of the policy model. Weaker models, such as Qwen2-VL, exhibit modest improvements, whereas the more powerful GPT-4o achieves significantly larger gains. This suggests that a model's capacity to effectively extract and integrate visual information is contingent upon its inherent capabilities.  Stronger models, possessing superior reasoning abilities, are able to leverage multi-modal cues more efficiently, resulting in greater performance improvements under inference-time scaling.
\section{Discussion}

% \paragraph{The necessity and applications of multi-modal thought.}

% From our point of view, multi-modal thought can convey richer features, which may not be easily and completely expressed in text. We also find a similar opinion in the latest work~\cite{Towards_reasoning_era}. These features (e.g., auxiliary lines for geometric problems in Figure~\ref{fig:cot-mcot}) can help generate better subsequent actions, thereby achieving better performance. While the high token cost currently hinders its scalability in human-interactive applications, a promising direction lies in leveraging multi-modal thought to automatically generate high-quality reasoning data. This data can be used to enhance the performance of LVLMs, building on the success of text-based reasoning approaches such as Mulberry~\cite{yao2024mulberry}, ALPHALLM~\cite{tts-mcts7}, and ALPHALLM-CPL~\cite{wang2024towards}.

\paragraph{The necessity of multi-modal thought.}
In this study, we empirically demonstrate that multi-modal thought possesses unique potential in representing cognitive processes that transcend the limitations of text alone. We argue that this approach can convey richer cognitive features, a perspective that aligns with the opinions from \citet{Towards_reasoning_era}.
A representative example, illustrated in Figure~\ref{fig:cot-mcot}, involves geometric reasoning. Here, visual-spatial elements (e.g., the use of auxiliary lines) provide critical insights that are inherently difficult to articulate through textual descriptions, thereby enhancing conceptual clarity. By integrating such multi-modal thought, the policy can engage in more informed decision-making processes, systematically improving problem-solving performance.

\paragraph{Reducing token costs for multi-modal thought to achieve efficient inference.}

Our analysis of token computations (Figure~\ref{fig:cost}) demonstrates that multi-modal thought, while effective, incurs substantial token costs.  This underscores the need for efficient compression techniques specifically designed for processing visual information in LVLMs. While prior works~\cite{cost1, cost2, cost3} have explored visual token compression, they have largely overlooked the unique challenges of multi-modal thought reasoning.  Thus developing methods for efficient visual information handling in this context becomes a key direction for future research.

\paragraph{Constructing stronger multi-modal verifiers for better performance.}

Our error analysis (\textsection\ref{subsec:error_analysis}) highlights the critical role of the verifier in inference-time scaling, where its effectiveness directly influences error reduction and performance gains. However, research on multi-modal verifiers~\cite{mm_reward2, mm_reward3, mm_reward4, mm_reward1} remains in its early stages, limiting the potential of multi-modal thought. Advancing more robust and accurate verifiers is therefore crucial for fully unlocking the benefits of inference-time scaling.

\paragraph{Exploring inference-time scaling for the chain of any-modal thought reasoning.}

Recent works have extended CoT reasoning beyond text and images to more modalities, such as audio~\cite{speech_cot} and video~\cite{video_cot}. However, the effectiveness of inference-time scaling for these modalities remains unexplored. While our findings confirm the benefits of scaling for text-image reasoning, whether these advantages extend to other modality combinations remains an open question.
% \subfile{sections/6.Related_works}
\section{Conclusion}

In this work, we present a systematic investigation of inference-time scaling with multi-modal thought.
We conduct comprehensive experiments across 10 challenging tasks spanning diverse domains, investigating both sampling-based and tree search-based scaling methods.
We demonstrate that multi-modal thought consistently outperform text-only reasoning, achieving progressively stronger performance with increased computational budgets.
This empirical evidence positions multi-modal thought scaling as a promising direction for enhancing complex reasoning capabilities in multi-modal scenarios.
However, we notice that performance gains are accompanied by higher computational costs and depend heavily on the effectiveness of the verifier, thus hindering real-world deployment.
We hope this work will serve as a springboard for studying inference-time scaling with multi-modal thought, thereby advancing multi-modal reasoning.

\section*{Limitations}

Due to the inability of current LVLMs to generate fine-grained images directly, we rely on generating image operation codes as a substitute, preventing a fully end-to-end multi-modal thought process. Additionally, we do not explore smaller-scale models, such as 7B, as their limited visual processing, instruction-following, and code generation abilities make them unsuitable for objectively evaluating multi-modal thought under inference-time scaling. Another limitation is the exclusion of training-based methods that require long CoT data, as current LVLMs struggle to generate high-quality long reasoning chains over multi-modal datasets, making data construction a significant challenge. Lastly, our study focuses solely on text and image modalities, leaving the generalizability of our findings to other modalities, such as audio and video, an open question.

\section*{Acknowledgments}
The project was supported by 
National Key R\&D Program of China (No. 2022ZD0160501), 
Natural Science Foundation of Fujian Province of China (No. 2024J011001),
and
the Public Technology Service Platform Project of Xiamen (No.3502Z20231043).
We also thank the reviewers for their insightful comments.

\bibliography{custom}

\appendix

% \usepackage{graphicx}
\begin{table*}[t]
\centering
\resizebox{0.8\textwidth}{!}{%
\begin{tabular}{c|lc}
\toprule
\textbf{Dataset}           & \multicolumn{1}{c}{\textbf{Description}}            & \textbf{Size}  \\ \midrule

Maxflow         & Find the maximum flow through a network.                & 128  \\ \midrule

Isomorphism       & Determine structural equivalence of two graphs.                                          & 128  \\\midrule
Connectivity      &  Determine path existence between two graph vertices.           & 128  \\\midrule

Convexity         &    Determine whether a function is convex or concave.   & 256  \\\midrule

Parity            & Determine if a function is even, odd, or neither.         & 384  \\\midrule

Winner ID         & Determine chess game outcome from the final board state.        & 257  \\\midrule

$V^*$ Bench       & Questions about small items in an image.         & 238  \\\midrule

MMVP              &  Visual questions designed to reveal CLIP-based LM shortcomings.        & 300  \\\midrule

BLINK             &   Visual perception tasks challenging for multi-modal LMs.         & 728  \\ \bottomrule
\end{tabular}%
}
\caption{The detailed introduction of the datasets in our experiments.}
\label{tab:static}
\end{table*}

\section{Implementation Details of Multi-modal Thought}
\label{sec:app_vsk}

Recognizing the limitations of current generative models in fine-grained drawing, we leverage Visual Sketchpad~\cite{vsk} to enable multi-modal reasoning.  Visual Sketchpad empowers LVLMs to generate sketches as intermediate reasoning artifacts, boosting problem-solving capabilities. It interprets multi-modal queries, devises sketching strategies, synthesizes programs for sketch generation, and analyzes the resulting sketches to formulate responses.

 Specifically, at each step $i$, the model outputs a textual thought $\mathbf{t}_i$ and a visual operation $o_i$:

\begin{equation}
    \mathbf{t}_i, o_i = \mathcal{M}\big(\mathbf{q}, \mathbf{I}, \mathbf{s}_{1:i-1}\big),
\end{equation}
where $\mathbf{o}_i$ is executable code specifying visual operations.

A code executor $\mathcal{G}$ then updates the visual contexts by applying $o_i$ to the previous images $\mathbf{I}$:

\begin{equation}
    \mathbf{I}' = \mathcal{G}\big(\mathbf{I}, o_i\big),
\end{equation}

\begin{equation}
    \mathbf{I} = \mathbf{I} \cup \mathbf{I}',
\end{equation}
where $\mathbf{I}'$ represents the updated images. Thus the next reasoning step $\mathbf{s}_i = \{\mathbf{t}_i, o_i, \mathbf{I}'\}$. 

Visual Sketchpad integrates various tools for sketch generation. For geometric and mathematical tasks, it uses Python libraries like matplotlib and networkx. For VQA, it incorporates specialized vision models and visual manipulations, including:

\begin{itemize}
\item \textbf{Detection:} Use Grounding-DINO~\cite{Groundingdino} to detect and label objects in images based on textual queries.
\item \textbf{Segmentation:} Generate segmented images with labeled masks by employing SegmentAnything~\cite{SegmentAnything} and Semantic-SAM~\cite{SemanticSam}.
\item \textbf{Depth Estimation:} Leverage DepthAnything~\cite{DepthAnything} to produce depth maps from input images.
\item \textbf{Visual Search:} Implements a sliding window approach to locate small objects based on textual queries.
\item \textbf{Image Manipulation:} Include zoom/crop (region of interest extraction) and image overlay (alpha blending).
\end{itemize}

\section{Prompts for Verifiers}
\label{sec:app_verifier_inst}

We use the following prompts for two kinds of verifiers:

\begin{itemize}
    \item \textbf{Classification-Based:} \textit{Verify the reasoning process above and provide the final judgment of 'yes' or 'no' on whether the reasoning is valid at last after “Final Decision:”.}
    \item \textbf{Regression-Based:} \textit{Verify the reasoning process above and provide a validation score from 0 (worst) to 1.0 (best) at last after “Final Score:”.}
\end{itemize}

\begin{figure}[t]
  \includegraphics[width=\columnwidth]{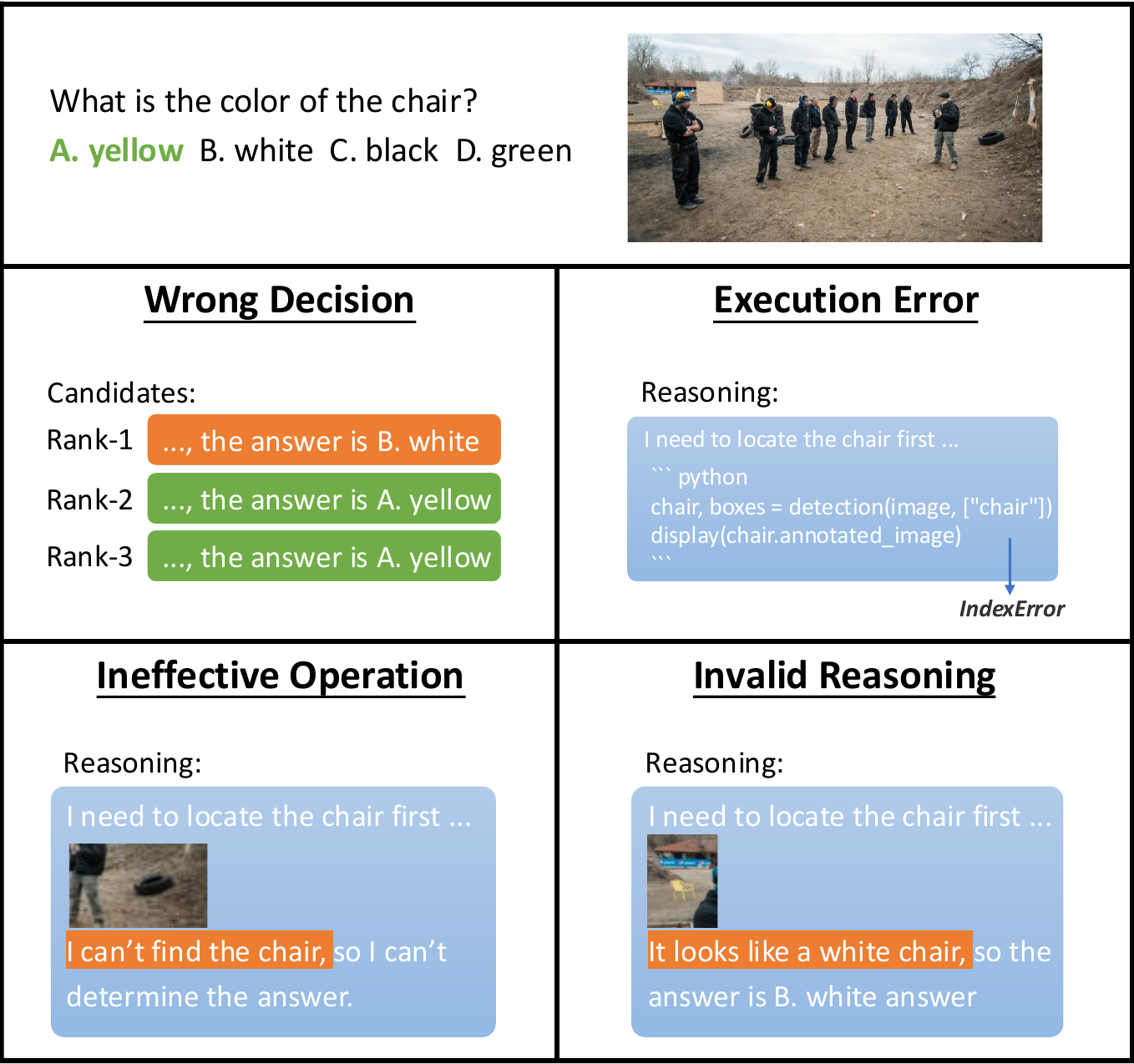}
  \caption{Examples of each error category.}
  \label{fig:error_type}
\end{figure}

\section{Datasets}
\label{sec:app_dataset}

The detailed introduction of datasets in our experiments are presented in Table~\ref{tab:static}.

\section{Baseline Introduction}
\label{sec:app_baselines}

Here we introduce the latest multi-modal reasoning baselines in our experiments:

\begin{itemize}
    \item \textbf{Multimodal-CoT}~\cite{mcot}: A multi-modal extension of CoT that concatenates the reasoning process with the problem context for answering.
    \item \textbf{DDCoT}~\cite{ddcot}: A method that decouples reasoning and recognition, and then integrate visual capabilities into the reasoning process.
    \item \textbf{CCoT}~\cite{ccot} A framework that generates a scene graph before answering.
\end{itemize}

\section{Error Category Explanation}
\label{sec:app_error}

We define four categories of error of multi-modal thought under inference-time scaling methods as follows:

\begin{itemize}
\item \textbf{Wrong Decision:} The correct answer is present within the candidate reasoning chains but is not selected. This indicates an error in the decision-making process.
\item \textbf{Execution Error:} The code generated during the reasoning process contains bugs, preventing the reasoning process from proceeding correctly.
\item \textbf{Ineffective Operation:} The reasoning process is hampered by ineffective operations that produce unhelpful or misleading visual information.
\item \textbf{Invalid Reasoning:} The error stems from a flawed reasoning process, leading to an incorrect conclusion.
\end{itemize}

Examples of each error category are provided in Figure~\ref{fig:error_type}.

\section{Additional Experiments}
\label{sec:add_exp}

\subsection{\texorpdfstring{Test Results on M$^{3}$CoT}{Test Results on M3CoT}}
\label{subsec:m3cot}
To validate the effectiveness on more challenging datasets, we conduct evaluations on M$^3$CoT~\cite{chen2024m3cot} datasets using GPT-4o, as presented in Table~\ref{tab:m3cot}.
These results demonstrate that the scaling of multi-modal thought continues to yield performance improvements in more complex scenarios.

\begin{table}[t]
\centering
\resizebox{0.55\columnwidth}{!}{%
\begin{tabular}{lc}
\toprule
Methods           & M$^3$CoT \\ \midrule
Directly          & 64.4  \\
CoT               & 65.0  \\
\ \ +Self-Consistency & 68.2  \\
\ \ +Best-of-N        & 68.8  \\
Visual Sketchpad  & 66.4  \\
\ \ +Self-Consistency & 69.0  \\
\ \ +Best-of-N        & 70.4  \\ \bottomrule
\end{tabular}%
}
\caption{Performance on M$^3$CoT using GPT-4o.}
\label{tab:m3cot}
\end{table}

\subsection{The Influence of the Image Quality}

To investigate the impact of image quality in the reasoning chain, we conduct additional experiments in which the resolution of intermediate images is artificially degraded by a scaling factor $\alpha$ (where a larger $\alpha$ corresponds to lower resolution and poorer quality). The results are reported in Table~\ref{tab:quality}.

These results show that reduced image quality (i.e., increased $\alpha$) leads to a clear degradation in reasoning performance and we also observe a qualitative reduction in inference latency with lower-resolution images. This suggests a potential trade-off between reasoning accuracy and computational efficiency, which merits further exploration.

\begin{table}[]
\centering
\resizebox{0.5\columnwidth}{!}{%
\begin{tabular}{lc}
\toprule
Methods          & BLINK \\ \midrule
Visual Sketchpad & 58.2  \\
+$\alpha$=0.2       & 57.1  \\
+$\alpha$=0.4       & 54.8  \\
+$\alpha$=0.6       & 47.4  \\ \bottomrule
\end{tabular}%
}
\caption{Performance on BLINK using multi-modal thought with varying intermediate image resolution scale factors ($\alpha$).}
\label{tab:quality}
\end{table}

\end{document}